\documentclass{article}

\usepackage{arxiv}
\usepackage[utf8]{inputenc}
\usepackage[T1]{fontenc}
\usepackage{hyperref}
\usepackage{url}
\usepackage{booktabs}
\usepackage{amsfonts}
\usepackage{nicefrac}
\usepackage{microtype}
\usepackage{lipsum}
\usepackage{graphicx}
\usepackage{natbib}
\usepackage{doi}
\usepackage{multirow}
\usepackage{tikz}
\usepackage{tabularx}
\usetikzlibrary{arrows.meta}

\usepackage{caption}
\usepackage{algorithm}
\usepackage{algorithmic}

\usepackage{newfloat}
\usepackage{listings}
\DeclareCaptionStyle{ruled}{labelfont=normalfont,labelsep=colon,strut=off}
\floatstyle{ruled}
\newfloat{listing}{tb}{lst}{}
\floatname{listing}{Listing}

\usepackage{booktabs}
\usepackage{amsmath}
\usepackage{cleveref}
\usepackage{subcaption}

\usepackage{amsmath}

\usepackage[utf8]{inputenc}

\usepackage{minted}
\title{The Real Cost of Context: Predicting Compute Time and Energy Consumption in LLM Prefill and Decode Phases}

\author{\href{https://orcid.org/0000-0001-6294-2915}{\includegraphics[scale=0.06]{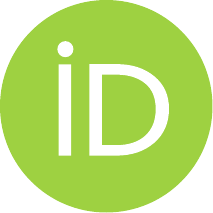}\hspace{1mm}Lars Krupp}\\
	RPTU University Kaiserslautern-Landau,\\German Research Center for Artificial Intelligence\\
	\texttt{kruppl@rptu.de} \\
	\And
	\href{https://orcid.org/0000-0003-2643-4504}{\includegraphics[scale=0.06]{orcid.pdf}\hspace{1mm}Daniel Geißler} \\
	RPTU University Kaiserslautern-Landau,\\German Research Center for Artificial Intelligence\\
	\texttt{daniel.geissler@dfki.de} \\
    \And
	\href{https://orcid.org/0000-0001-6647-4215}{\includegraphics[scale=0.06]{orcid.pdf}\hspace{1mm}Francisco M. Calatrava-Nicolas} \\
	Örebro University\\
	\texttt{francisco.calatrava-nicolas@oru.se} \\
    \And
    \href{https://orcid.org/0009-0006-0038-1183}{\includegraphics[scale=0.06]{orcid.pdf}\hspace{1mm}Stefan Gerd Fritsch} \\
	German Research Center for Artificial Intelligence\\
    \texttt{stefan\_gerd.fritsch@dfki.de} \\
    \And
    \href{https://orcid.org/0000-0003-0320-6656}{\includegraphics[scale=0.06]{orcid.pdf}\hspace{1mm}Paul Lukowicz} \\
	RPTU University Kaiserslautern-Landau,\\German Research Center for Artificial Intelligence\\
    \texttt{paul.lukowicz@dfki.de} \\
    \And
	\href{https://orcid.org/0000-0002-0698-7470}{\includegraphics[scale=0.06]{orcid.pdf}\hspace{1mm}Jakob Karolus} \\
	RPTU University Kaiserslautern-Landau,\\German Research Center for Artificial Intelligence\\
	\texttt{jakob.karolus@dfki.de} \\
}

\renewcommand{\shorttitle}{The Real Cost of Context}

\usepackage{xcolor}
\begin{document}
\maketitle
\begin{abstract}

The integration of large language models (LLMs) into agentic architectures has substantially increased token consumption. Yet, existing energy estimates focus primarily on generated output tokens and often underestimate energy consumption for the large input contexts characteristic of agentic workflows.
To address this gap, we systematically profile compute time and energy consumption locally across a wide range of prefill (input) and decode (output) token lengths. Using generalized linear models, we show that prefill compute time and energy exhibit both linear and quadratic scaling with input length.
We then use our local compute time models to estimate energy consumption for API deployments, where direct energy measurements are unavailable, by mapping the closest local configuration, thus providing a workload equivalent for the API’s energy consumption. Our work revises existing models for estimating LLM inference time and energy supporting more sustainable and informed deployment decisions in the era of agentic AI.

\end{abstract}




\section{Introduction}

Large Language Models (LLMs) have evolved from primarily supporting isolated conversational interactions to becoming foundational components of increasingly complex applications. As their adoption and usage continue to grow, accurately estimating their energy consumption is becoming an important operational and environmental concern~\cite{luccioni2023estimating,samsi2023words}.  For open-weight models deployed on accessible hardware, energy consumption can be measured through software telemetry or external power meters~\cite{argerich2024measuring,husom2025sustainable}.

At the same time, modern LLM applications increasingly process large amounts of contextual information. This trend is particularly visible in multi-turn and agentic workflows, where conversation histories, system instructions, retrieved documents, tool outputs, and intermediate reasoning steps may be repeatedly included in the model input. Systems such as OpenClaw\footnote{\url{https://github.com/openclaw/openclaw}}
illustrate how complex agentic workflows can substantially increase input token consumption when solving long-running, multi-step tasks~\cite{bai2026ai}. Likewise, this challenge applies to any LLM application that processes long input contexts. Despite this development, existing energy estimation approaches largely focus on output tokens only and use short and fixed input lengths~\cite{argerich2024measuring,wilkins2024offline,alizadeh2027cost}. Consequently, the significant impact of input length on compute time as well as energy consumption remains inadequately modeled, particularly for large-context workloads.

We address this gap by systematically profiling LLM compute time and energy consumption across a grid of input and output token lengths, independently varying the tokens processed during prefill (input) and those produced during the decode (output) stage.

Based on these empirical observations, we fitted generalized linear models (GLMs) on locally collected compute time and energy data, capturing the contributions of input and output tokens. We show that the number of input tokens has both linear as well as quadratic effects on compute time and energy consumption. These results demonstrate that estimates based predominantly on generated tokens~\cite{ecologits, argerich2024measuring} may fail to capture an important component of inference cost for long-context requests.

The fitted GLMs provide a compact and explainable representation of the LLM's compute time and energy behavior. For each local hardware and software configuration, we fit a GLM pair, one predicting compute time and the other predicting energy consumption for input--output configurations.

In contrast, proprietary models accessed through commercial APIs expose little information about their underlying hardware, serving configuration, or power consumption, making their energy footprint considerably more difficult to estimate~\cite{ecologits,krupp2026taking}.

Because energy consumption depends on both execution time and power draw, compute time predictions alone are insufficient, particularly for proprietary APIs whose underlying hardware and power characteristics are not observable. Following the local-equivalence approach proposed by Krupp et al.~\cite{krupp2026taking}, we compare the API compute time GLM with the local compute time GLMs and select the configuration whose behavior is most similar over the evaluated workloads. We refer to this configuration as a \emph{workload equivalent} and use its paired energy GLM to obtain a locally grounded estimate of API energy consumption.

Our work presents an empirically grounded and explainable modeling approach for estimating LLM compute times and, through locally measured power characteristics, their associated energy consumption. Our method is suitable for both local and API-accessed LLMs and requires no access to the proprietary serving infrastructure. It is particularly relevant to context-intensive applications, including increasingly common multi-turn and agentic workflows.

We outline our primary contributions as follows:

\begin{enumerate}
    \item We systematically characterize and quantify the influence of input token length on LLM compute time, with particular focus on the large-context requests increasingly prevalent in modern LLM applications.

    \item We develop generalized linear models (GLMs) that model LLM compute time and energy consumption from local measurements while providing good interpretability through their direct correspondence to the algorithmic characteristics of LLM inference.

    \item We use compute time to identify a local workload equivalent for an API deployment and use its paired energy model to estimate API energy consumption under the power characteristics of the selected local configuration.

\end{enumerate}

\section{Related Work}


\subsection{Modeling Inference Time and Energy}

The environmental impact of large-scale machine learning systems has received increasing attention in recent years. Early work
primarily framed this issue around the cost of training large neural models, showing that gains in model accuracy often come with substantial financial and environmental costs, and proposing that model comparison should account for both predictive performance and computational resources~\cite{strubell2019energy}.
Further work showed the dependence of carbon estimation on model architecture, accelerator type, datacenter efficiency, and geographic energy mix, making estimation difficult when these factors are not reported~\cite{patterson2021carbon}.

With the widespread deployment of LLMs, research on the environmental impact of machine learning has increasingly expanded from training to inference.
The lifecycle assessment of BLOOM included its deployment phase, measuring energy consumption over approximately 18 days of serving real API requests and observing continued consumption even during periods of minimal demand~\cite{luccioni2023estimating}.
Further work investigated the impact of model size, GPU generations, workloads, and distributed multi-GPU configurations on inference energy~\cite{samsi2023words}, and evaluated the trade-offs among quantization, output accuracy, latency, and energy when running LLMs on a Raspberry Pi~\cite{husom2025sustainable}.
Others attributed per-prompt energy to active accelerators, host CPU and memory, provisioned idle capacity, and datacenter overhead using internal telemetry from the Gemini Apps serving fleet~\cite{elsworth2025measuring}.
These studies show that the energy attributed to inference is determined not only by the model and the request, but also by the hardware and infrastructure through which the request is served.

While these studies focus primarily on measuring and attributing inference energy, complementary work has examined the serving mechanisms that determine how computional resources are used during inference.
LLM inference was shown to have distinct power phases, with a compute-intensive prefill phase and a longer decode phase with different utilization characteristics~\cite{patel2024characterizing}, allowing their separation across different hardware resources~\cite{patel2024splitwise}. The influence of input and output sequence length on inference energy was further investigated by Cavagna et al.~\cite{pizzini2026sweetspot}, who proposed an analytical model accounting for computational and memory access costs and identified non-linear relationships between sequence lengths and energy efficiency. At the application level, Matinpour et al.~\cite{matinpour2026evaluation} investigated the impact of orchestration frameworks, foundation models, and deployment constraints, demonstrating that framework selection affects system performance beyond the choice of the underlying model. They also observed increasing end-to-end latency with larger context windows, although their analysis does not separately characterise the contributions of the prefill and decoding phases.

Finally, detailed descriptions of production inference have been made, describing separate prefill and decoding deployments together with GPU allocation, pipeline parallelism, expert parallelism, and redundant expert placement~\cite{liu2024deepseek}.

Together, these studies show that inference energy depends on distinct prefill and decoding costs as well as on serving decisions such as batching, hardware allocation, and parallelism.

Existing inference-energy evaluations have also focused predominantly on relatively short inputs (see \Cref{tab:related_work}).

Studies used prompts with lengths of 8~\cite{argerich2024measuring}, 115--1360~\cite{alizadeh2027cost}, and  approximately 64---4096~\cite{pizzini2026sweetspot} tokens per request. This range is extended to 50,000 input tokens with fixed output length, but only in the context of locally deployed models running on known hardware~\cite{caravaca2025prompts}.


\begin{table}[h!]

    \centering
    \footnotesize
    \begin{tabular}{p{3cm}p{1.5cm}p{5.5cm}p{4.5cm}}
    \toprule
    Estimation Approach & Input Token Range (Aprox.) & Datasets Used & Modelling of Input Token Contribution \\
    \midrule
    Carbon Footprint of BLOOM~\cite{luccioni2023estimating} & n/r & Live API traffic (230k requests) & Not modelled; energy related to request count only \\ 
    Measuring and Improving the Energy Efficiency of LLM Inference~\cite{argerich2024measuring} & 8 & 
    The Pile~\cite{gao2020pile}, Arxiv~\cite{arxiv_org_submitters_2024}
    & 
    Implicitly included in total inference energy; not modelled separately \\ 
    EcoLogits~\cite{ecologits} & 
    1--54835* 
    & 
    ML.ENERGY~\cite{chung2026ml} (arena-human-preference-140k, gpqa-diamond, sourcegraph-fim) & Not modelled: GPU energy $\propto$ output tokens; constant time-to-first-token in host energy only \\ 
    AI Energy Score~\cite{aiEnergyScore} & 2--18083* & WikiText~\cite{merity2016pointer}, OSCAR~\cite{suarez2020monolingual}, UltraChat~\cite{ding2023enhancing}, CNN/DailyMail~\cite{nallapati2016abstractive}, SAMSum~\cite{gliwa2019samsum}, ArXiv, SQuAD v2~\cite{rajpurkar2018know}, XTreme~\cite{hu2020xtreme}, SquadShifts~\cite{miller2020effect}, IMDB~\cite{maas2011learning}, Tweet Eval~\cite{barbieri2020tweeteval}, Yelp, STS-B~\cite{cer2017semeval} & Measured per phase (prefill, decode), summed per query; not modelled \\
    Towards Sustainable NLP~\cite{poddar2025towards} & 100--400 & CoLA~\cite{warstadt-etal-2019-neural}, SST-2~\cite{socher-etal-2013-recursive}, MNLI~\cite{williams-etal-2018-broad}, BoolQ~\cite{clark-etal-2019-boolq}, COPA~\cite{roemmele2011choice},ReCoRD~\cite{zhang2018record}, SQuAD v2~\cite{rajpurkar2018know}, CNN/DailyMail~\cite{nallapati-etal-2016-abstractive}, SAMSum~\cite{gliwa-etal-2019-samsum}, VAX-Stance~\cite{poddar2022winds}, CAVES~\cite{poddar2022caves} & Measured per sample; empirically linear (controlled sweep 100--400) \\
    Offline Energy-Optimal LLM Serving~\cite{wilkins2024offline} & 8--2048 & Synthetic sweep; Alpaca~\cite{alpaca} 
    & Linear energy regression with input–output interaction, fitted separately for each LLM\\ 
    
    SweetSpot~\cite{pizzini2026sweetspot} & 64--4096 & Synthetic fixed-length requests & Analytical model with $I^2$, $I$, $I O$, $O$, $O^2$ terms \\ 
    TokenPowerBench~\cite{niu2026tokenpowerbench} & 0--10k & Alpaca~\cite{alpaca}, LongBench~\cite{bai2024longbench}, or custom prompts & Measured per phase; bucketed context lengths (0--2k, 2--5k, 5k--10k \\ 
    How Hungry is AI?~\cite{jegham2025hungry} & 100--10k & Artificial Analysis prompt sizes (100 / 1k / 10k) & Implicitly included in total inference energy; not seperately modelled \\ 
    Energy Considerations of LLM Inference~\cite{fernandez2025energy} & 50--32k & WMT-14~\cite{bojar2014findings}, WikiText-2~\cite{merity2016pointer}, IMDB~\cite{maas2011learning}, CNN/DailyMail~\cite{nallapati2016abstractive}, BurstGPT~\cite{wang2025burstgpt}, Azure chat/code traces~\cite{stojkovic2025dynamollm}; PG-19~\cite{rae2019compressive} sweep & Measured prefill and decode separately; no explicit modelling\\ 
    From Prompts to Power~\cite{caravaca2025prompts} & 10--50k & synthetic promts &  Not explicitly stated \\ 
    The cost of AI-assisted coding~\cite{alizadeh2027cost} &115--1360& HumanEvalPack(Python)~\cite{muennighoff2024octopack} & modeled linearly, and as interaction term with active parameter count; reported as having negligible impact\\ 

    \midrule
    \textbf{Ours} & 10k--200k & Synthetic (random nonce, filler sentences, fixed instruction) & Fitted model with linear ($I$), quadratic ($I^2$) and $I{\times}O$ terms \\
    \bottomrule
    \end{tabular}
    \caption{Prior LLM inference energy estimation approaches, sorted by the maximum input length they cover across all associated datasets and sweeps. $I$/$O$: input/output tokens; n/r: not reported. *The EcoLogits ($\text{median}~=~44$, $\bar{x}=~291$, $s=848$) and AI Energy Score ($\text{median} =  121$, $\bar x = ~229$, $s=~ 489$) datasets are highly unbalanced.}
    \label{tab:related_work}
\end{table}

\subsection{Estimating Energy for API Deployments}

However, commercial APIs expose only request-level information such as token numbers and response time, leaving the underlying infrastructure and power consumption hidden. This limits the applicability of estimators requiring direct measurements or detailed system knowledge and motivates approaches based on observable request information~\cite{krupp2026promoting}.

EcoLogits~\cite{ecologits}, the closest approach to ours,
models GPU energy using the number of output tokens, active model parameters, and an assumed batch size, while host-system energy is estimated from the request duration, inferred hardware requirements, and datacenter overhead. However, input token length is not explicitly represented in its energy or compute time models, so the additional prefill cost associated with longer prompts is not separately captured.

Consequently, the timing behavior of proprietary API-based LLMs over substantially longer contexts, and its implications for energy estimation, remains insufficiently characterized. We address this gap by systematically varying input length up to approximately 200,000 tokens while controlling output length and measuring its effect on request duration, compute time, and energy consumption under explicit hardware and power assumptions.

\section{Characterizing Local Compute Time and Energy Consumption}
\label{sec:local_characterization}

Existing estimation approaches often emphasize output generation, treating input token processing as negligible~\cite{ecologits,jegham2025hungry,kim2026toward,argerich2024measuring}. This assumption becomes questionable for modern long-context and agentic workloads. We therefore first characterize the effects of input and output length on locally measured compute time and energy consumption. We evaluate our approach on Mistral-Small-2603, a Mixture-of-Experts (MoE) model with 119B total parameters, of which 6B are active. Its size allows it to fit on two H200 GPUs, making it a candidate for local deployment and testing.

We deploy the model locally under several hardware configurations, and batch sizes 1,2,4,8, and 16 ($2^0$--$2^4$) fitting pairs of compute time and energy GLMs from measured compute time and GPU power consumption. We examine how compute time and energy consumption depend on input and output length. \Cref{sec:api_workload} applies these models to API energy estimation. A complete overview of our method is depicted in \Cref{fig:method_workflow}.

\begin{figure*}[t]
  \centering
  \begingroup
  \definecolor{workflowInk}{HTML}{203040}
  \definecolor{workflowMuted}{HTML}{526575}
  \definecolor{workflowBenchmark}{HTML}{F3F6F8}
  \definecolor{workflowLocal}{HTML}{EDF3FA}
  \definecolor{workflowLocalArrow}{HTML}{356CA5}
  \definecolor{workflowAPI}{HTML}{F5F0F9}
  \definecolor{workflowAPIAccent}{HTML}{8255A2}
  \definecolor{workflowProblem}{HTML}{9F3038}
  \definecolor{workflowAlert}{HTML}{FBECEE}
  \definecolor{workflowMatch}{HTML}{EAF5F1}
  \definecolor{workflowMatchAccent}{HTML}{267567}
  \definecolor{workflowEnergy}{HTML}{FFF1E2}
  \definecolor{workflowEnergyAccent}{HTML}{975222}
  \newcommand{\workflowfont}[1]{%
    \fontencoding{T1}\fontfamily{phv}\fontsize{#1}{#1}\selectfont}
  \resizebox{\linewidth}{!}{%
  \begin{tikzpicture}[
    x=24mm, y=24mm,
    every node/.style={
      anchor=west, inner sep=0pt, text=workflowInk,
      font=\workflowfont{11.8}
    },
    panel/.style={rounded corners=8pt},
    heading/.style={font=\workflowfont{14}\bfseries},
    model/.style={anchor=center,font=\workflowfont{11.2}\bfseries},
    arrow/.style={
      -{Triangle[length=5.6pt,width=6.4pt]}, line width=1.4pt,
      shorten >=1.5pt, shorten <=1.5pt
    }
  ]
    \path[use as bounding box] (0,0) rectangle (12,8.17);

    \fill[panel,workflowBenchmark] (.1,7.27) rectangle (11.9,8.07);
    \node[heading] at (.34,7.82) {Experiment Configuration};
    \node[font=\workflowfont{12.1}] at (.34,7.51)
      {Mistral-Small-2603 \enspace\textperiodcentered\enspace
       Input 10k--200k tokens (step 10k) \enspace\textperiodcentered\enspace
       Output 500--2000 tokens (step 500)};

    \fill[panel,workflowLocal] (.1,3.60) rectangle (6.2,6.98);
    \node[heading] at (.34,6.69) {Local profiling (\Cref{sec:local_characterization})};
    \node[text=workflowMuted,font=\workflowfont{10.4}] at (.34,6.43)
      {Local Configurations};
    \node at (.34,6.03) {H100 \texttimes{} 4 / TP=4};
    \node at (.34,5.73) {H200 \texttimes{} 2 / TP=2};
    \node at (.34,5.43) {H200 \texttimes{} 4 / TP=4};
    \node[anchor=center,font=\workflowfont{23}] at (2.92,5.73) {\texttimes};
    \node[font=\workflowfont{11.8}\bfseries] at (3.31,6.03) {Batch sizes};
    \node at (3.31,5.69) {1 / 2 / 4 / 8 / 16};
    \node[model] at (3.15,5.19) {(Total of 15 configurations)};
    \draw[workflowLocalArrow,line width=.9pt] (3.15,5.09) -- (3.15,4.76);
    \draw[arrow,workflowLocalArrow,shorten <=0pt] (3.15,4.76) -| (1.505,4.33);
    \draw[arrow,workflowLocalArrow,shorten <=0pt] (3.15,4.76) -| (4.60,4.33);

    \draw[workflowLocalArrow,line width=.9pt] (.48,4.25) circle[radius=4pt];
    \draw[workflowLocalArrow,line width=.9pt,line cap=round]
      (.48,4.25) ++(0,2.3pt) -- (.48,4.25) -- ++(1.8pt,-.9pt);
    \node[font=\workflowfont{10.2}] at (.61,4.25) {Compute time measurements};
    \fill[workflowEnergyAccent]
      (3.55,4.32) -- (3.49,4.24) -- (3.55,4.24) --
      (3.52,4.18) -- (3.62,4.27) -- (3.56,4.27) -- cycle;
    \node[font=\workflowfont{10.2}] at (3.70,4.25) {GPU energy measurements};

    \fill[panel,white] (.34,3.73) rectangle (2.85,4.07);
    \fill[panel,white] (3.06,3.73) rectangle (5.96,4.07);
    \node[model] at (1.595,3.90) {15 compute time GLMs};
    \node[model] at (4.51,3.90) {15 energy GLMs};

    \fill[panel,workflowAPI] (7.02,3.60) rectangle (11.9,6.98);
    \node[heading,text=workflowAPIAccent] at (7.27,6.69) {API profiling (\Cref{sec:api_time})};
    \fill[panel,workflowAlert] (7.27,5.55) rectangle (11.65,6.32);
    \draw[workflowProblem,line width=1.25pt,line join=round]
      (7.62,6.12) -- (7.43,5.78) -- (7.81,5.78) -- cycle;
    \draw[workflowProblem,line width=1.2pt,line cap=round]
      (7.62,6.01) -- (7.62,5.89);
    \fill[workflowProblem] (7.62,5.83) circle[radius=1pt];
    \node[align=left,text=workflowProblem,font=\workflowfont{11.5}\bfseries]
      at (7.97,5.935) {Direct energy measurements\\are not possible.};
    \draw[arrow,workflowAPIAccent] (9.465,5.44) -- (9.465,4.40);
    \draw[workflowAPIAccent,line width=.9pt] (8.43,4.25) circle[radius=4pt];
    \draw[workflowAPIAccent,line width=.9pt,line cap=round]
      (8.43,4.25) ++(0,2.3pt) -- (8.43,4.25) -- ++(1.8pt,-.9pt);
    \node[font=\workflowfont{10.2}] at (8.56,4.25) {Compute time measurements};
    \fill[panel,white] (7.72,3.73) rectangle (11.21,4.07);
    \node[model] at (9.465,3.90) {API compute time GLM};

    \draw[arrow,workflowLocalArrow] (1.595,3.59) -- (1.595,3.17);
    \draw[arrow,workflowAPIAccent] (9.465,3.59) -- (9.465,3.17);
    \fill[panel,workflowMatch] (.1,1.22) rectangle (11.9,3.15);
    \node[heading,text=workflowMatchAccent] at (.34,2.91)
      {Match compute time behavior (\Cref{sec:select_workload})};
    \node[model] at (2.05,2.52) {15 local compute time GLMs};
    \foreach \column in {0,...,4} {
      \foreach \row in {0,...,2} {
        \fill[workflowLocalArrow]
          ({1.53+.26*\column},{1.90+.18*\row}) circle[radius=2.6pt];
      }
    }
    \node[model] at (9.465,2.52) {API compute time GLM};
    \fill[workflowAPIAccent] (9.465,2.08) circle[radius=4.2pt];
    \fill[panel,white] (4.63,1.87) rectangle (7.37,2.29);
    \node[model,text=workflowMatchAccent] at (6,2.08) {Workload Equivalent};
    \draw[arrow,workflowLocalArrow] (2.82,2.08) -- (4.55,2.08);
    \draw[arrow,workflowAPIAccent] (9.29,2.08) -- (7.45,2.08);
    \node[anchor=center,font=\workflowfont{11.5}] at (6,1.49)
      {Based on root mean squared error (RMSE): H200 \texttimes{} 2
       \textperiodcentered{} TP=2 \textperiodcentered{} batch=1};

    \draw[arrow,workflowEnergyAccent] (6,1.21) -- (6,.94);
    \fill[panel,workflowEnergy] (.1,.10) rectangle (11.9,.92);
    \node[heading,anchor=center,text=workflowEnergyAccent] at (6,.66) {Estimate API energy (\Cref{sec:api_energy})};
    \node[anchor=center] at (6,.34)
      {Use the energy GLM paired with the selected workload equivalent.};
  \end{tikzpicture}%
  }
  \endgroup
  \caption{Overview of our workload equivalent method. The shared token
  grid supports the comparison of local and API compute time models. The
  energy estimatation then uses the paired energy model of the closest local
  configuration.}
  \label{fig:method_workflow}
\end{figure*}

\subsection{Benchmark Construction and Experimental Design}
\label{sec:benchmark}

Crucially, our benchmark only aims at generating a prompt of a fixed input token length, while ensuring a fixed output length.
To do so, we constructed each prompt out of three parts. Each prompt starts with a randomly generated 32bit nonce (number used once) to inhibit prefix caching, costing approximately 64 tokens. This ensures that the computations required in the prefill phase have to be executed each time and results are not retrieved from the cache.
The second part consists of random sentences that can be extended until a defined input token length is reached, allowing the input token length to be scaled to arbitrarily large numbers.
Finally, we added a prompt (310 tokens) that generates a large number of output tokens. To control the output token length, we used the max output token parameter~\cite{kwon2023efficient,mistral_chat_api}, which stops the decoding step once the desired output length is reached. The complete prompt is provided in the supplementary material.
Using this setup, the number of input tokens of each prompt and the output tokens of the LLM's response can be finely controlled. Since LLMs are statistical models, a desired output token length cannot be guaranteed. We safeguard against this case through reruns.

Existing work commonly quantifies LLM inference energy, focusing on tokens generated during the decode phase~\cite{ecologits},
leaving the impact of the prefill phase on time and energy consumption underexplored. However, we argue that the prefill phase is becoming increasingly important as modern LLM-based systems process progressively longer input contexts.

To assess the importance of the prefill and decode phases on energy consumption, we start by investigating their impact on compute time (as energy is power over time). We set up our benchmark as a coarse grid-search, testing input token lengths between 10,000 and 200,000 with a step size of 10,000.  For each input length, we test four output lengths: 500, 1,000, 1,500, and 2,000 tokens. We repeat each input--output combination 80 times, resulting in 80 unique combinations (20 prompt lengths $I$ $\times$ 4 output lengths $O$). These combinations cover a broad region of the prefill space, allowing us to identify trends over a large input token range while simultaneously considering an output token range comparable to those examined in previous approaches, providing us with a more holistic understanding of compute time $t_{compute}(I, O)$.

If the general assumption in the field
holds, the results of our experiments will show that increasing input token numbers has little to no impact on the overall compute time.
On the other hand, we expect to see more than a linear increase in compute time since the number of prefill operations scales quadratically with the number of input tokens~\cite{vaswani2017attention}.

We acknowledge the use of generative AI models (Claude, OpenAI, Google) to support coding, data analysis, and the prototyping of figures.

\subsection{Local Configurations and Energy Measurement}
\label{sec:local_measurement}

To obtain the local compute time and energy measurements, we ran the benchmark described in \Cref{sec:benchmark} with Mistral-Small-2603 hosted using vLLM~\cite{kwon2023efficient}. We enabled chunked prefilling (\texttt{max\_\allowbreak num\_\allowbreak batched\_\allowbreak tokens=16384}) in three different configurations (H100 4 GPU, TP=4; H200 2 GPU, TP=2; H200 4 GPU, TP=4) and five different batch sizes ($2^0$--$2^4$) using tensor parallelism (TP) restricting GPU memory utilization to 85\% and max model length to 220,000 tokens. All local experiments used 16-bit floating-point (FP16) precision. In addition to recording the compute time, we used nvidia-smi to capture the power and GPU utilization for all GPUs.

The resulting energy data was postprocessed individually for each input--output configuration and divided into sections corresponding to individual prompts\footnote{Each input--output configuration consists of 80 prompts (cf.~\Cref{sec:benchmark}).}.

To calculate how much energy was consumed in a section, we use numerical integration. We define $P_x$ as the total power consumed by all GPUs for running the LLM during that section at timestamp $t_x$.
The result represents the energy $E(I, O)$ consumed by one prefill and one decode phase for one specific prompt with $I$ input tokens that resulted in $O$ output tokens.
\begin{equation}
    E(I,O) = \sum_{x=1}^{n-1}\frac{P_x+P_{x+1}}{2} \times (t_{x+1}-t_x)
\end{equation}

\subsection{Modeling Compute Time and Energy}
\label{sec:local_models}

To characterize compute time and energy consumption, we fitted statistical models relating both outcomes to the respective numbers of input and output tokens. The models should be interpretable and reflect the computational structure of transformer inference. We therefore favored statistical models over approaches such as deep neural networks, which may achieve low prediction errors but typically require more training data and provide limited insight into the modeled relationships. Specifically, we employ generalized linear models (GLMs) with a quasipoisson error structure and an identity link function.

\begin{enumerate}
    \item Compute time measurement variance increases with compute time. The \textbf{Quasipoisson error structure} allows variance to scale with mean compute time: $\text{Var}(Y_i) = \theta \mu_i$, where $\theta$ is the dispersion parameter.
    \item We omitted the default log-link function in favor of an \textbf{identity link} ($g(\mu_i) = \mu_i$). The identity link directly maps to the additive nature of the hardware operations on the GPU (first prefill and then decode). Additionally, as the decode phase is sequential and limited by memory bandwidth, each generated output token incurs a flat, additive execution cost on the GPU.
\end{enumerate}

To fit our models, we used the data points collected in our coarse grid search to fit one compute time and one energy consumption GLM per local configuration. This allows us to characterize the relationship between input and output token numbers and both measured outcomes.

We filtered the local datasets using the standardized deviance residuals of the preliminary GLM fit. Observations with absolute deviance residuals
of 3 or greater were removed to eliminate background anomalies and prevent outlier masking during final parameter estimation.

For all compute time and energy GLMs, we initially fitted a model containing $I, I^2, O, I \times O, I^2 \times O$ and $O^2$. The $I^2 \times O$ and $O^2$ terms were found to be statistically insignificant. To preserve model simplicity and avoid multicollinearity, we pruned these terms. Residual diagnostics confirmed that the standardized residuals followed the normal diagonal, while leverage analysis verified no remaining influential data points.

After removing these terms, the final fitted mean response takes the form
\begin{equation}
    \widehat{y}(I,O)=\beta_0+\beta_1 I+\beta_2 I^2+\beta_3 O+\beta_4 I O.
    \label{eq:paired_glm}
\end{equation}
The coefficients are estimated separately for compute time and energy consumption. The common model structure allows their dependence on the same input--output workload to be compared.

\subsection{Effects of Input and Output Length}
\label{sec:local_effects}

\begin{figure*}[h]
    \centering
    \includegraphics[width=\linewidth]{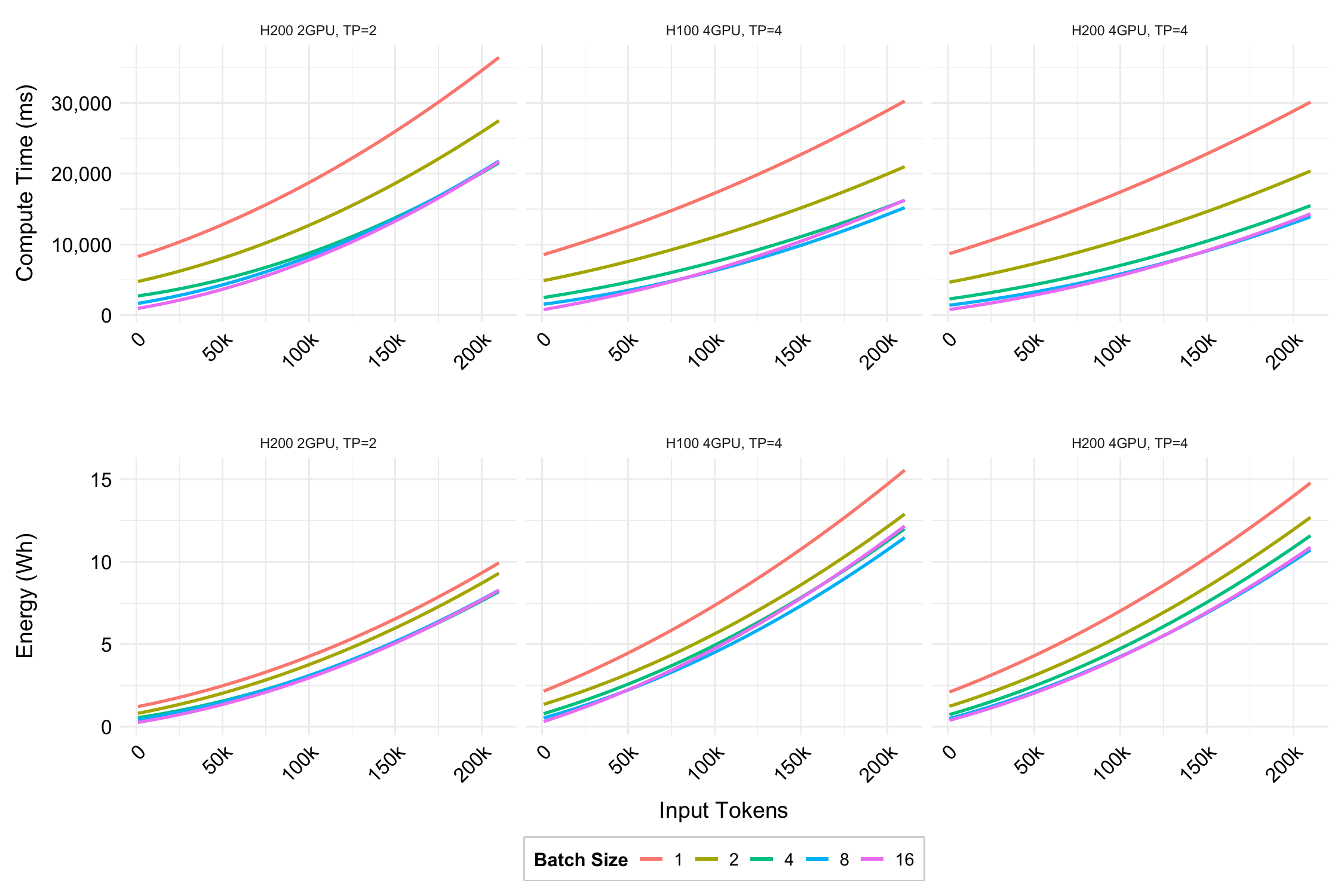}
    \caption{GLM Prediction Curves for compute time (ms) and energy (Wh) between 0 and 200,000 input tokens and 2000 output tokens for batch sizes 1, 2, 4, 8, and 16 sorted by hardware configuration.}
    \label{fig:facets_all}
\end{figure*}

Using the compute time and energy data obtained for the different hardware configurations, we fitted GLMs that take the number of input tokens and output tokens to predict both outcomes. We find both linear as well as quadratic influence of input tokens on compute time and energy consumption, as depicted in \Cref{fig:facets_all} for a fixed ouput size. The influence of output tokens and the interaction of input tokens and output tokens remain linear in both sets of models. These results show analogous workload dependence for compute time and energy consumption. We also evaluated GPU utilization and measured a median of 100\% across configurations, indicating good use of the available hardware resources.

The local time and energy models therefore contain both $I$ and $I^2$ terms, together with $O$ and $I O$. For a fixed input length, the predicted contribution of output tokens is linear. For a fixed output length, the predicted response includes both linear and quadratic contributions from input length.

The estimated coefficients can be interpreted in terms of the computational structure of transformer inference. The intercept represents baseline system overhead, including scheduling and other contributions captured by the positive response floor. The linear and quadratic input terms describe the cost of processing the prompt, with the quadratic component consistent with the $O(N^2)$ complexity of self-attention for extended context windows~\cite{vaswani2017attention}. Prefill is highly parallelized and compute intensive, whereas autoregressive decoding is sequential and constrained by memory bandwidth~\cite{patel2024characterizing}. The linear output term describes the incremental cost of generating tokens, while the interaction term captures how that cost changes with input context length.



This behavior is consistent with the theoretical stucture of transformer inference. Prefill and decoding process the same input--output workload, while their different power profiles determine how elapsed time contributes to consumed energy~\cite{vaswani2017attention,patel2024characterizing}. Together with the local measurements, this provides a basis for using compute time behavior to select a local energy model for a comparable workload. The coefficients of the time and energy models can differ because power draw varies across phases and configurations.




\section{Estimating API Energy Through Workload Equivalents}
\label{sec:api_workload}

\begin{figure*}[h]
    \centering
    \includegraphics[width=\linewidth]{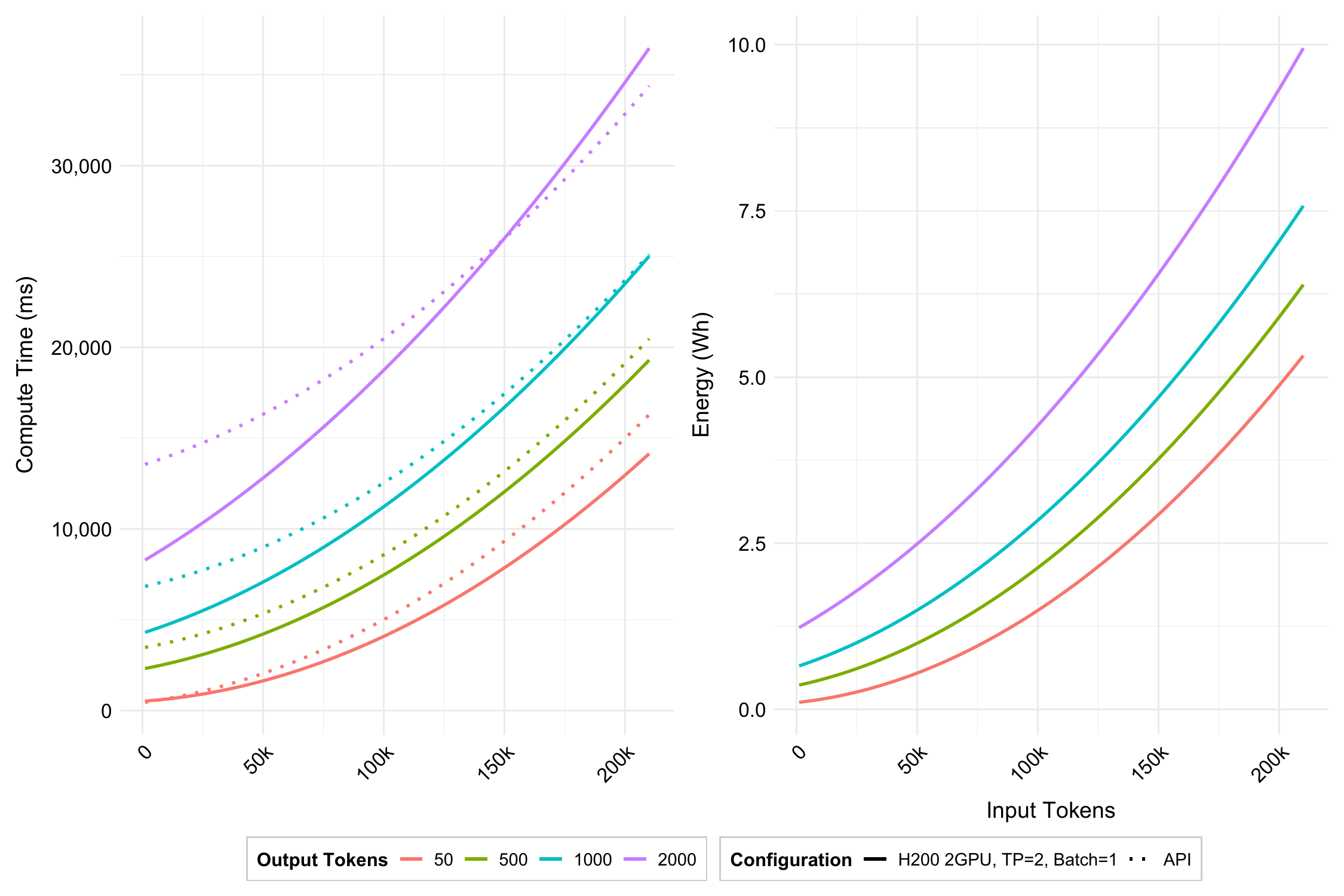}
    \caption{Compute time and energy GLM curves for the workload equivalent (H200 2GPU, TP=2, Batch=1) and the API GLM.}
    \label{fig:best_fit}
\end{figure*}

The energy consumption of an API-based LLM cannot be measured directly since the underlying serving infrastructure is inaccessible. We now use compute time, which can be estimated from API response times, to connect the API deployment with locally hosted configurations for which both compute time and energy consumption can be measured. For each local configuration, the paired GLMs describe how time and energy respond to the input--output workload. We identify the local compute time GLM that most closely matches the API compute time GLM and use its paired energy GLM to estimate energy consumption (see \Cref{fig:best_fit}).

We call the selected configuration a \emph{workload equivalent} for the API over the evaluated input--output range. The local findings motivate the assumption that similar compute time behavior across these workloads can support a useful energy proxy. The transferred energy estimate is conditional on the power characteristics of the selected local configuration.

\subsection{Modeling API Compute Time}
\label{sec:api_time}

To better isolate the time attributable to LLM computation time, we developed a benchmark specifically for time and energy estimation that incorporated a network latency estimation procedure, thereby reducing the influence of noise associated with API calls.

We evaluated Mistral-Small-2603\footnote{\url{https://docs.mistral.ai/models/model-cards/mistral-small-4-0-26-03}} through Mistral's API using the same controlled input and output lengths as in \Cref{sec:benchmark}.

We can now formulate the total LLM response time as follows:
\begin{equation}
\begin{aligned}
     t_{total}(I,O) = t_{compute}(I,O) + t_{net} + t_{other}\\
\end{aligned}
\end{equation}

The noise introduced by the network conditions consists of network latency $t_{net}$ and other $t_{other}$ additional serving overhead. To reduce this noise, we ping the server every second and subtract this estimated network latency from the total compute time. We therefore define the network-adjusted response time as:
\begin{equation}
\begin{aligned}
    t_{\mathrm{adj}}(I,O)
    &= t_{\mathrm{total}}(I,O) - t_{\mathrm{net}} \\
    &= t_{\mathrm{compute}}(I,O) + t_{\mathrm{other}} \\
    &\approx t_{\mathrm{compute}}(I,O) \\
    &\qquad \text{when } t_{\mathrm{compute}}(I,O) \gg t_{\mathrm{other}}
\end{aligned}
\end{equation}

As we are conducting our experiment with large token lengths, $t_{other}$ is negligible as $t_{compute}>>t_{other}$. Nonetheless, our regression approach (see \Cref{sec:local_models}) is able to implicitly capture this overhead as well.

We fitted the API GLM using the shared fitting and model-selection procedure described in \Cref{sec:local_models}. Before fitting, outliers were removed from the API data using Tukey's 1.5$\times$ interquartile range (IQR) rule to reduce noise. We then applied the same standardized deviance residual threshold used for the local models. 
Removing anomalous response times helps reduce the influence of transient network traffic and server workload on the estimated relationship between token counts and compute time.

The API coefficient estimates follow the interpretation described in \Cref{sec:local_effects}. Our API GLM estimates a baseline system overhead of $\beta_0 \approx 91.97\text{ ms}$. While statistically non-significant, retaining this parameter represents the positive compute time floor associated with scheduling and other overheads captured by $t_{other}$.

For input tokens, the estimated linear component is $\beta_1=0.018\text{ ms/token}$ and the quadratic component is $\beta_2=2.70\times10^{-7}\text{ ms/token}^2$. The output coefficient is $\beta_3=6.71\text{ ms/token}$, while the input--output interaction is $\beta_4=1.23\times10^{-5}\text{ ms/token}^2$. At input length $I$, the predicted incremental compute time per output token is therefore $\beta_3+\beta_4 I$.

Overall, the API GLM model demonstrates high explanatory power, accounting for $98.8\%$ of the total deviance.



\begin{table}[t]
\centering
\footnotesize
\begin{tabular}{lrrr}
\toprule
& \textbf{API} & \multicolumn{2}{c}{\textbf{H200 (2GPU, TP=2, batch=1)}} \\
\cmidrule(lr){2-2} \cmidrule(lr){3-4}
\textbf{Predictor} & \textbf{Time (ms)} & \textbf{Time (ms)} & \textbf{Energy (Wh)} \\
\midrule
$\beta_0$: Intercept    & $91.97$               & $309.0$               &  $7.66 \times10^{- 2}$  \\
$\beta_1$: $I$          & $1.81 \times 10^{-2}$ & $0.738\times 10^{-2}$ &  $3.47 \times10^{- 6}$  \\
$\beta_2$: $I^2$        & $2.70 \times 10^{-7}$ & $2.65\times 10^{-7}$  &  $9.97 \times10^{-11}$  \\
$\beta_3$: $O$          & $6.71$                & $3.94 $               &  $5.67 \times10^{- 4}$  \\
$\beta_4$: $I \times O$ & $1.23 \times 10^{-5}$ & $3.57 \times 10^{-5}$ &  $8.59 \times10^{- 9}$  \\
\bottomrule
\end{tabular}
\caption{Quasipoisson GLM parameter estimates for time of Mistral-Small-2603 accessed through Mistrals API (time) and the local workload equivalent (time,energy).}

\label{tab:final_results}
\end{table}

\subsection{Identifying a Workload Equivalent}
\label{sec:select_workload}

We compare the compute time GLM of each of the 15 local configurations $k$ with the API GLM over the shared grid $\mathcal{G}=\mathcal{I}\times\mathcal{O}$. To reduce the influence of constant overhead, we set the intercept of each GLM to zero. At grid point $g$, each GLM then predicts
\begin{equation}
    \hat y_g (I,O)=\beta_1 I_g+\beta_2 I_g^2+\beta_3 O_g+\beta_4 I_g O_g.
\end{equation}
We define the compute time distance $D_k$ as the root mean squared difference between the local and API predictions over the shared grid.
\begin{equation}
\begin{aligned}
    d_{k,g} &= \hat y_{k,g}^{\mathrm{local}}-\hat y_g^{\mathrm{API}},\\
    D_k &= \sqrt{\frac{1}{|\mathcal{G}|}\sum_{g\in\mathcal{G}}d_{k,g}^2}.
\end{aligned}
\label{eq:workload_rmse}
\end{equation}
We select the local configuration with the smallest distance as the workload equivalent, $k^*=\arg\min_k D_k$.

\begin{table}[t]

    \centering
    \begin{tabular}{crrr|r}
    \toprule
     Config & Batch & $R^2$ (time) & $R^2$ (energy) & $D_k$ (ms) \\
     \midrule
H200 2GPU, TP=2 & 1  & 0.995 &0.995& 1870. \\
H100 4GPU, TP=4 & 1  & 0.994 &0.994& 3367. \\
H200 4GPU, TP=4 & 1  & 0.996 &0.995& 3450. \\
H200 2GPU, TP=2 & 2  & 0.996 &0.997&  4970. \\
H100 4GPU, TP=4 & 2  & 0.996 &0.996&  7108. \\
H200 4GPU, TP=4 & 2  & 0.995 &0.995&  7395. \\
H200 2GPU, TP=2 & 4  & 0.994 &0.994&  7516. \\
H200 2GPU, TP=2 & 8  & 0.994 &0.996&  7707. \\
H200 2GPU, TP=2 & 16 & 0.995 &0.996&  7949. \\
H100 4GPU, TP=4 & 4  & 0.994 &0.995&  9204. \\
H200 4GPU, TP=4 & 4  & 0.993 &0.994&  9583. \\
H100 4GPU, TP=4 & 16 & 0.991 &0.993&  9828. \\
H100 4GPU, TP=4 & 8  & 0.990 &0.992& 10085. \\
H200 4GPU, TP=4 & 8  & 0.990 &0.991& 10399. \\
H200 4GPU, TP=4 & 16 & 0.993 &0.995& 10420. \\
\bottomrule
    \end{tabular}
    \caption{Configurations with goodness of fit values for compute time and energy GLMs ($R^2$). The fit of the API compute time GLM is $R^2$ = 0.987. Additionally, $D_k$ is listed as the distance of each configuration to the API compute time GLM.}
    \label{tab:RMSE}
\end{table}

The minimum distance is achieved by the configuration H200 2GPU, TP=2 with batch size 1 (\Cref{tab:RMSE}). We use this configuration as the API's workload equivalent over the evaluated token grid. \Cref{tab:final_results} depicts the GLM coefficients for the API compute time as well as the coefficient for its closest workload equivalent (compute time and energy models).

\subsection{Estimating API Energy Consumption}
\label{sec:api_energy}

The local time and energy models show analogous dependence on the input--output workload. The API compute time model exhibits the same linear and quadratic input contributions, motivating the assumption that its energy consumption also reflects these workload-dependent costs. Under this assumption, we use the energy GLM paired with the selected workload equivalent to estimate the energy consumed by the API-accessed LLM.

For input length $I$ and output length $O$, the API energy estimate is the predicted energy $\widehat{E}_{k^*}(I,O)$ of the local configuration $k^*$ selected by the compute time RMSE. This retains the workload dependence measured locally while adopting the power characteristics of that configuration.

We therefore use this configuration to estimate the energy consumed by the model hosted on Mistral's servers, while acknowledging that closer approximations may be found by evaluating more configurations.

\subsection{Comparing Energy Estimates}

\Cref{tab:related_work} groups prior approaches by how they treat input tokens. Some omit an explicit input term, and some model input length linearly or quadratically. EcoLogits illustrates the first approach in its GPU energy estimate~\cite{ecologits}. At fixed output length, the model of Wilkins et al.~\cite{wilkins2024offline} is linear in input length, and SweetSpot~\cite{pizzini2026sweetspot} includes a quadratic term.

Our energy GLM includes both linear ($I$) and quadratic ($I^2$) input terms, as well as a linear output term ($O$) and an interaction ($I\times O$). The interaction allows the linear contribution of input length to vary with output length, while $I^2$ captures the curvature in energy as prompts grow.

\begin{figure*}[t]
    \centering
    \includegraphics[width=\linewidth]{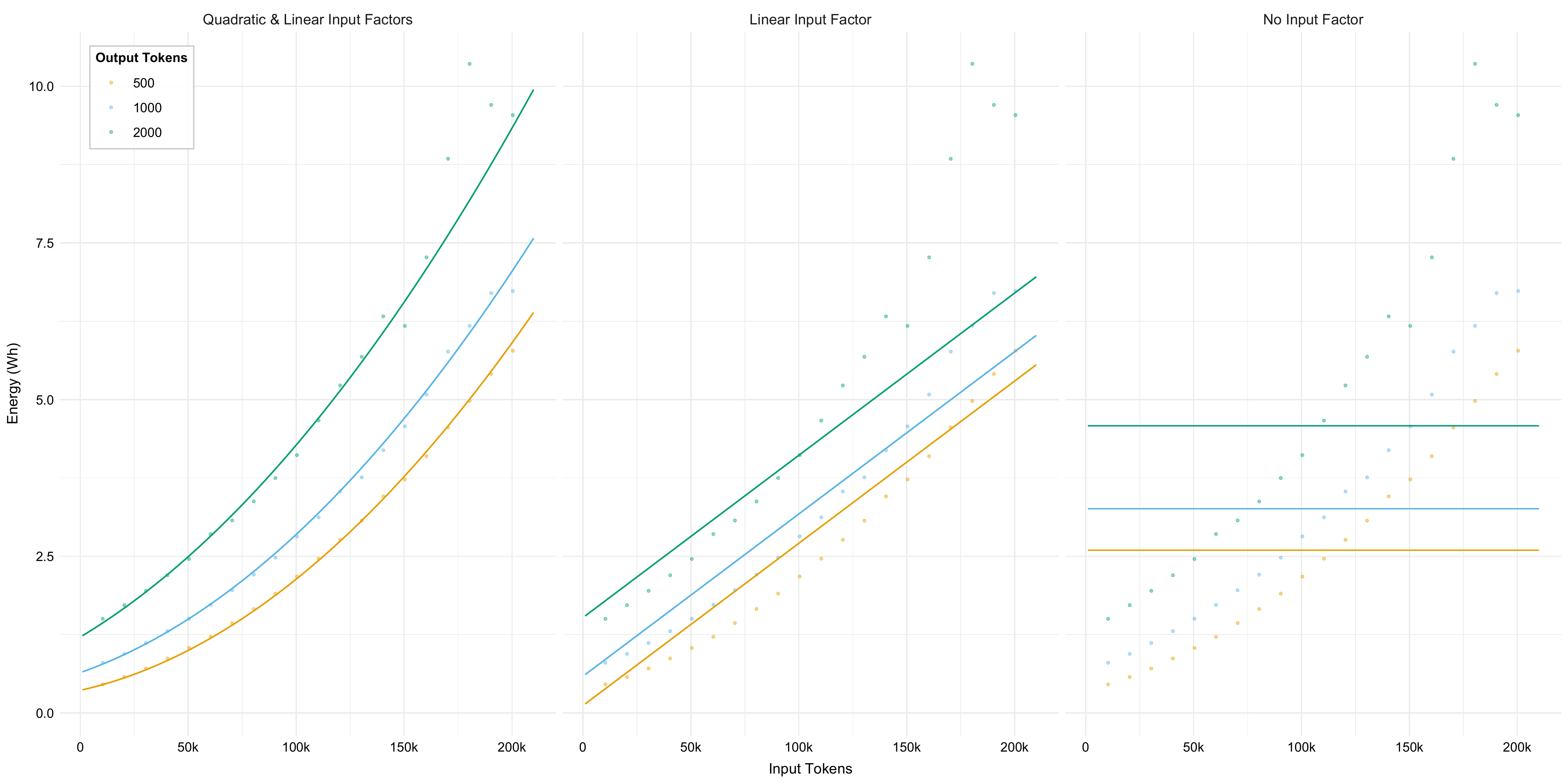}
    \caption{Comparison between different model fits on our local energy measurements of the workload equivalent configuration (H200 2GPU, TP=2, Batch=1), visualized for 500, 1000, and 2000 output tokens. Left side: our model with a quadratic and linear component of $I$. Middle: only linear component of $I$. Right: No factor for $I$, essentially omitting the impact of input token size.}
    \label{fig:energy_model_forms_panels}
\end{figure*}

\Cref{fig:energy_model_forms_panels} shows how each model form follows the local measurements at 500, 1000 and 2000 output tokens. Our model (on the far left) uses a quadratic and linear component of $I$ offering the closest match to our observed energy measurements. The solely linear fit (middle) approximates energy over parts of the shorter input range and falls below our model beyond roughly 150,000 input tokens. Completely omitting any factors for $I$ results in a constant energy prediction for a given output token size. At the longest measured input length, all simpler approaches underestimate the measured mean.

As our empirical results have confirmed, prefill time depends on input length, and assuming it to be constant or linear can lead to inaccurate energy estimates, especially at large context sizes. Current approaches to measuring and estimating inference energy therefore need to be reevaluated using workloads with large numbers of input tokens.

\section{Discussion}

The results of our experiments show the relevance of input tokens when estimating the compute time and energy consumption of LLMs. Our approach uses GLM pairings to describe their dependence on the input--output workload. These local models then support the selection of a workload equivalent for API energy estimation.

\subsection{Prefill Time Must Not Be Neglected}

Prior works of both empirical~\cite{patel2024characterizing} and theoretical nature~\cite{vaswani2017attention} have already shown that the prefill phase scales non-linearly with input tokens, while the decode phase scales linearly.

However, the role of this information in LLM energy estimation remains insufficiently studied.
Our results show that the impact of the prefill phase on overall compute time becomes significant for large numbers of input tokens and therefore has to be considered for LLM energy estimation. We believe that improper benchmarks for which the number of input tokens rarely exceeds 4000 (see \Cref{tab:related_work}) have resulted in previous work underestimating the impact of the prefill phase on energy consumption. An example is GPQA~\cite{rein2023gpqa}, which uses complex, but short, questions. This benchmark is reasonable to use for evaluating LLM capabilities, but not when evaluating energy consumption.

While currently established estimation approaches can be accurate for small input lengths, for larger input lengths, these approaches will systematically underestimate the compute time and consumed energy. In an era where large context sizes are standard, e.g., in AI agents and complex RAG systems, we need to consider energy estimation approaches capable of properly modeling all contributing factors.

\subsection{Scaling Our Approach Through Workload Equivalents}

The fitted GLMs provide an interpretable representation of how compute time and energy vary with input and output length. Especially the better explainability and direct alignment with the transformer architecture, compared to purely data-driven methods, offer a robust approach for real-world benchmarking.

The API application depends on identifying a local workload equivalent whose compute time behavior closely matches that of the API-accessed model. Additional measurements across hardware, software, and batch configurations can expand the available comparisons and may improve the match over the workloads of interest.

While our approach has some distinct disadvantages, such as the cost associated with exploring this configuration space, the benefits are significant. Apart from enabling the estimation of the compute time and energy consumption for otherwise completely opaque models, the GLMs can support the development of more sustainable agentic AI systems. They can inform model selection, software, and hardware configurations based on expected input and output token lengths and their associated energy footprints.
As LLM context sizes continue to grow, knowing how compute time and energy consumption scale with it becomes increasingly important.

To expand the available GLM pairs, we further plan to involve the larger research community through hosting a website\footnote{[removed for review]} that aggregates the GLMs and data recordings of other models and configurations, including different batch sizes.

\subsection{Limitations}

\subsubsection{Coverage of Local Configurations and Workloads}

We were able to conduct our experiments only on a limited number of hardware and software configurations, due to limited computational resources. While an exact match with the API version of Mistral-Small-2603 was not among them, our empirical data showed the non-negligible impact of input length on energy consumption. The selected workload equivalent is the closest approximation among the evaluated configurations and over the evaluated token range.




\subsubsection{Energy Transfer and Measurement Scope}

The shared GLM structure and the local time--energy relationship support the use of a workload equivalent, but the absolute energy estimate depends on local power characteristics. Similar compute time behavior can occur with different hardware, power draw, or serving arrangements. The API energy estimates should therefore be interpreted as estimates under the selected local configuration's power characteristics, rather than as direct measurements of the provider's energy consumption.

Our direct measurements cover GPU energy. Applying a PUE factor accounts for an assumed datacenter overhead, while other components of the provider's energy use, including host-system energy and shared idle capacity, are not directly measured. In addition, the fitted model terms reflect the evaluated model and serving configurations. Other architectures or serving mechanisms may require a different functional form, which can be investigated using the same profiling approach.

\subsubsection{Other Attention Mechanisms Impacting the Generalizability of our Fitted GLMs}
We evaluated Mistral-Small-2603, which uses Multi-Head Latent Attention. Models with other attention mechanisms, such as sliding window attention~\cite{fu2025sliding} or linear attention~\cite{katharopoulos2020transformers}, may exhibit different relationships between input length, compute time, and energy consumption. As a result, the estimated effects of the $I$, $I^2$, and $I \times O$ terms may vary across large language models. However, the same profiling method and GLM fitting procedure can still be applied to measurements of other LLMs to determine which terms are supported by their data.

\section{Conclusion}

Our empirical findings demonstrate that the prefill phase significantly impacts the compute time and energy consumption of LLMs, making it a critical factor for accurate estimation, especially as context sizes grow and agentic AI systems expand.

To address this, we introduced a generalized linear model (GLM) approach that estimates compute time and energy consumption from local measurements.

Because direct energy measurements are unavailable for proprietary LLMs, we then use compute time to identify a local workload equivalent and use its paired energy model to estimate API energy consumption. This extends the local characterization to API deployments under the power assumptions of the selected configuration.

Ultimately, these findings offer a scalable and reproducible way to estimate energy consumption, advancing transparency for proprietary LLMs and highlighting the growing necessity of sustainable AI practices.
\bibliographystyle{unsrtnat}
\bibliography{bibliography}


\end{document}